\documentclass[10pt,twocolumn,letterpaper]{article}

\usepackage{icb}
\usepackage{times}
\usepackage[bookmarks=false]{hyperref}
\usepackage{epsfig}
\usepackage{graphicx}
\usepackage{amsmath}
\usepackage{amssymb}
\usepackage{booktabs}
\usepackage{gensymb}
\usepackage{multirow}
\usepackage[norule,symbol,perpage]{footmisc}

\icbfinalcopy 

\ificbfinal\fi

\begin{document}


\title{Deep Pixel-wise Binary Supervision for Face Presentation Attack Detection}

\author{Anjith George and S\'ebastien Marcel \\
Idiap Research Institute \\
Rue Marconi 19, CH - 1920, Martigny, Switzerland \\
{\tt\small  \{anjith.george, sebastien.marcel\}@idiap.ch  }
}

\maketitle
\thispagestyle{empty}

\begin{abstract}
    Face recognition has evolved as a prominent biometric authentication modality. However, vulnerability to presentation attacks curtails its reliable deployment. Automatic detection of presentation attacks is essential for secure use of face recognition technology in unattended scenarios. In this work, we introduce a Convolutional Neural Network (CNN) based framework for presentation attack detection, with deep pixel-wise supervision. The framework uses only frame level information making it suitable for deployment in smart devices with minimal computational and time overhead. We demonstrate the effectiveness of the proposed approach in public datasets for both intra as well as cross-dataset experiments. The proposed approach achieves an HTER of 0\% in Replay Mobile dataset and an ACER of 0.42\% in \textit{Protocol-1} of OULU dataset outperforming state of the art methods.
\end{abstract}


\section{Introduction}

Face recognition has evolved as a prominent biometric authentication modality. The ubiquitous nature of face recognition can be mainly attributed to the ease of use and non-intrusive data acquisition.
Many of the recent works have reported human level parity in face recognition \cite{learned2016labeled}. While there is an increased interest in face recognition for access control, vulnerability to presentation attacks (PA) (also known as spoofing) curtails its reliable deployment. Merely presenting printed images or videos to the biometric sensor could fool face recognition systems. Typical examples of presentation attacks are print, video replay, and 3D masks \cite{costa2016replay}, \cite{erdogmus2014spoofing}. For the reliable use of face recognition systems, it is important to have automatic methods for detection of such presentation attacks. 

\begin{figure}[t]
\centering
        \includegraphics[width=0.98\linewidth]{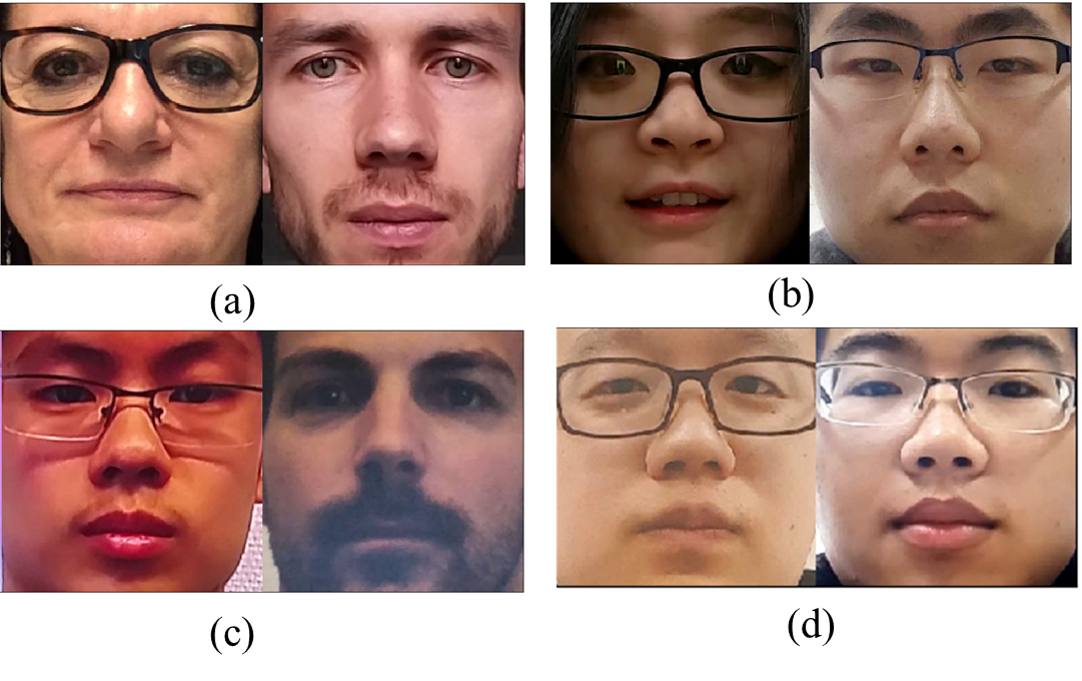}
        \caption{Figure showing cropped face images for \textit{bonafide} and presentation attacks in in Replay-Mobile \cite{costa2016replay} (a,c), and OULU-NPU \cite{boulkenafet2017oulu}datasets (b,d).}\label{fig:example_attacks}

\end{figure}

In literature, several authors have proposed presentation attack detection (PAD) algorithms for counteracting the presentation attack attempts \cite{ramachandra2017presentation}. Majority of the methods rely on the limitation of presentation attack instruments (PAI) and the quality degradation during recapture. Handcrafted features are specifically designed to utilize this degradation for PAD. Most of them use features extracted from color \cite{boulkenafet2015face}, texture \cite{maatta2011face},\cite{chingovska2012effectiveness}, motion \cite{anjos2011counter} and other liveliness cues.  

Recently several CNN based PAD algorithms have emerged \cite{gan20173d,yang2014learn,shao2017deep}, which learns the features for PAD without the requirement for designing handcrafted features. Even though CNNs trained end to end with binary PAD task achieved good intra dataset performance as compared to handcrafted feature based methods, they fail to generalize across databases and unseen attacks. Often, a limited amount of data is available to train CNNs from scratch which results in over fitting. It is possible that the network could learn the biases in a dataset since it learns explicitly from the given training data, resulting in poor cross database generalization. Some recent literature \cite{atoum2017face}, \cite{liu2018learning} have shown that the usage of auxiliary tasks such as depth supervision can improve the performance. The network learns to synthesize the depth map of the face region as an auxiliary task. Depth supervision requires synthesis of 3D shapes for every training sample. However, this synthesis can be avoided as the PAD task is not directly related to the depth estimation task. We show that deep pixel-wise binary supervision can be used for pixel-wise supervision obviating the requirement of depth synthesis.


Most of the PAD databases consists of videos (usually of 5 to 10 s  duration \cite{boulkenafet2017oulu},\cite{costa2016replay}). Usage of video instead of individual frames provides additional temporal information for PAD. However, in practical deployment scenarios such as mobile environments, the time available for acquisition and processing is limited. Algorithms achieving good performance using long video sequence may not be suitable for such deployment conditions where the decision needs to be made quickly. Hence, frame level PAD approaches are advantageous from the usability point of view since the PAD system can be integrated into a face recognition system with minimal time and computational overhead.

Motivated by the discussions above, we introduce a frame level CNN based framework for presentation attack detection. The proposed algorithm uses deep pixel-wise binary supervision for PAD (\textit{DeepPixBiS}). 

We demonstrate the effectiveness of the proposed approach in two public datasets namely Replay Mobile \cite{costa2016replay} and OULU \cite{boulkenafet2017oulu} databases. Sample images of the cropped face regions from both datasets are shown in Fig. \ref{fig:example_attacks}. Both intra and cross-dataset experiments are performed indicating the efficacy of the proposed method.

The main contributions of this paper are listed below, 

\begin{itemize}

\item A frame level CNN based framework is proposed for PAD, which is suitable for practical deployment scenarios since it requires only frames instead of videos.

\item Pixel-wise binary supervision is proposed which simplifies the problem and obviates the requirement for video samples and synthesis of depth maps. 


\item We show the efficacy of the proposed approach with the experiments in both intra as well as cross-database testing in recent publicly available datasets. 


\end{itemize}

Moreover, the results shown in this paper are fully reproducible. The protocols and source code to replicate experiments are made available \footnote{Source code available at: \url{https://gitlab.idiap.ch/bob/bob.paper.deep_pix_bis_pad.icb2019}}.

\section{Related work} 

Most of the literature in PAD can be broadly categorized as feature-based and CNN based methods. 

\subsection{Feature based methods}

Several methods have been proposed over the years for presentation attacks using handcrafted features. They can be further classified to methods based on color, texture, motion, liveliness cues and so on. Histogram features using color spaces  \cite{boulkenafet2015face}, local binary pattern \cite{maatta2011face}, \cite{chingovska2012effectiveness} and motion patterns \cite{anjos2011counter} have shown good performance in Replay Attack \cite{chingovska2012effectiveness} database. Image quality measures \cite{galbally2014image}, and image distortion analysis \cite{wen2015face} use the deterioration of the sample quality and artifacts in the re-capture as a cue for presentation attack detection. Most of these methods treat PAD as a binary classification problem which reduces its generalization capability in an unseen attack scenario ~\cite{nikisins2018effectiveness}. Nikisins \textit{et al}. \cite{nikisins2018effectiveness} proposed a framework for one class classification using one class Gaussian Mixture Models (GMM). Image Quality Measures (IQM) were used as the features in their work. For the experiments, they prepared an aggregated dataset combining Replay Attack \cite{chingovska2012effectiveness}, Replay Mobile \cite{costa2016replay}, and MSU-MFSD \cite{wen2015face} datasets.

Boulkenafet \textit{et al}. \cite{boulkenafet2017competition} compiled the results of a public competition to compare the generalization properties of the PAD algorithms in mobile environments. 
The OULU-NPU \cite{boulkenafet2017oulu} dataset was used to benchmark the algorithms. Several feature based methods and 
CNN based methods were compared in this competition. The GRADIANT system, which comprised of color, texture and motion information from different color spaces, was ranked first. In their approach, the dynamic information from the video is collapsed into a frame. Also, LBP features from small grids were concatenated to a feature vector. Feature selection was done using recursive feature selection, and SVM based classification is done on each feature vector, and sum fusion is used for the final PA score. 

\subsection{CNN based approaches}

Recently several authors have shown that CNN based methods achieve good performance in PAD.
Gan \textit{et al}. \cite{gan20173d} proposed a 3D-CNN  based approach which combines spatial and temporal features of the video for PAD. Yang \textit{et al}. \cite{yang2014learn} proposed a framework where the feature representation obtained from the trained CNN is used to train an SVM classifier and was used for the final PAD task. Li \textit{et al}. \cite{li2018learning} also proposed a 3D CNN architecture, where the Maximum Mean Discrepancy (MMD) distance among different domains is minimized to improve the generalization property. Shao \textit{et al}. \cite{shao2017deep} proposed a deep CNN based architecture for the detection of 3D mask attacks. In their approach, the subtle differences in facial dynamics captured using the CNN is used for PAD task. In each channel, feature maps obtained from the convolutional layers of a pretrained VGG network was used to extract features. They also estimated the optical flow in each channel and the dynamic texture was learned channel-wise. Their approach achieved an AUC (Area Under Curve) score of 99.99\% in 3DMAD \cite{erdogmus2014spoofing} dataset. However, this method is specifically tuned for the detection of 3D mask attacks, the performance in case of 2D attacks was not discussed.

 Li \textit{et al}. \cite{li2017face} proposed a part-based CNN model for PAD. In their method face region is divided into different parts and individual CNNs were trained for each part. Usage of patches increased the number of samples available for training the network. The network architecture used was based on VGG-face. The last layers from the models trained for each part were concatenated and used for SVM training, which in turn was used in the prediction stage. They obtained better results as compared to networks using the whole face at once.

Some of the main issues with CNN based methods are the limited amount of training data and poor generalization in unseen attacks and cross-database settings. To reduce these issues, some researchers have used auxiliary supervision in the training process. 

Atoum \textit{et al}. \cite{atoum2017face} proposed a two-stream CNN for 2D presentation attack detection combining the outputs from a patch-based CNN and depth map CNN.
An end to end CNN model was trained using random patches in the patch based part. A fully Convolutional network was trained to produce the depth map for \textit{bonafide} samples. A feature vector was computed from the depth map obtained from the depth CNN by finding the mean values in the $N \times N$ grid and used to train an SVM model. The final score from the system was generated by combining the scores from the patch and depth based systems. Though this method achieved good performance on intra dataset experiments in Replay attack \cite{chingovska2012effectiveness}, CASIA-FASD \cite{zhang2012face} and MSU-USSA \cite{patel2016secure} datasets, performance in challenging cross-database testing are not reported. 

Liu \textit{et al}. \cite{liu2018learning} proposed a CNN based approach which uses auxiliary supervision for PAD. They used a CNN-RNN model to compute the depth map with pixel-wise supervision as well as remote photoplethysmography (rPPG) signal for sequence wise supervision.  
In the testing phase, the estimated depth map and rPPG signal were used for PAD task. They showed that the addition of auxiliary tasks improves the generalization property. However, the higher accuracies reported uses temporal information, which requires more frames (hence more time). Such methods are not suitable for practical deployment scenarios from a usability point of view since a user would need to spend more time for authentication.

\subsection{Limitations}

From the discussions in the previous section, it can be seen that the PAD problem is far from solved and is very challenging. From the recent literature, it can be seen that CNN based methods outperform handcrafted feature-based methods. While training CNN for PAD, one of the main issues is the lack of availability of a sufficient amount of data for training a network from scratch. Further, the overfitting in the datasets with unacceptable cross-database performance is another issue. Some of the recent approaches require the fusion of multiple systems; which makes it complicated for deployment. Another limiting factor is the usage of video in many algorithms. 
In the mobile authentication scenario, the time available for the PAD decision is very short. Hence frame based PAD methods may be advantageous from the usability point of view.






\section{Proposed method}

\begin{figure}[ht]
\centering
        \includegraphics[width=0.98\linewidth]{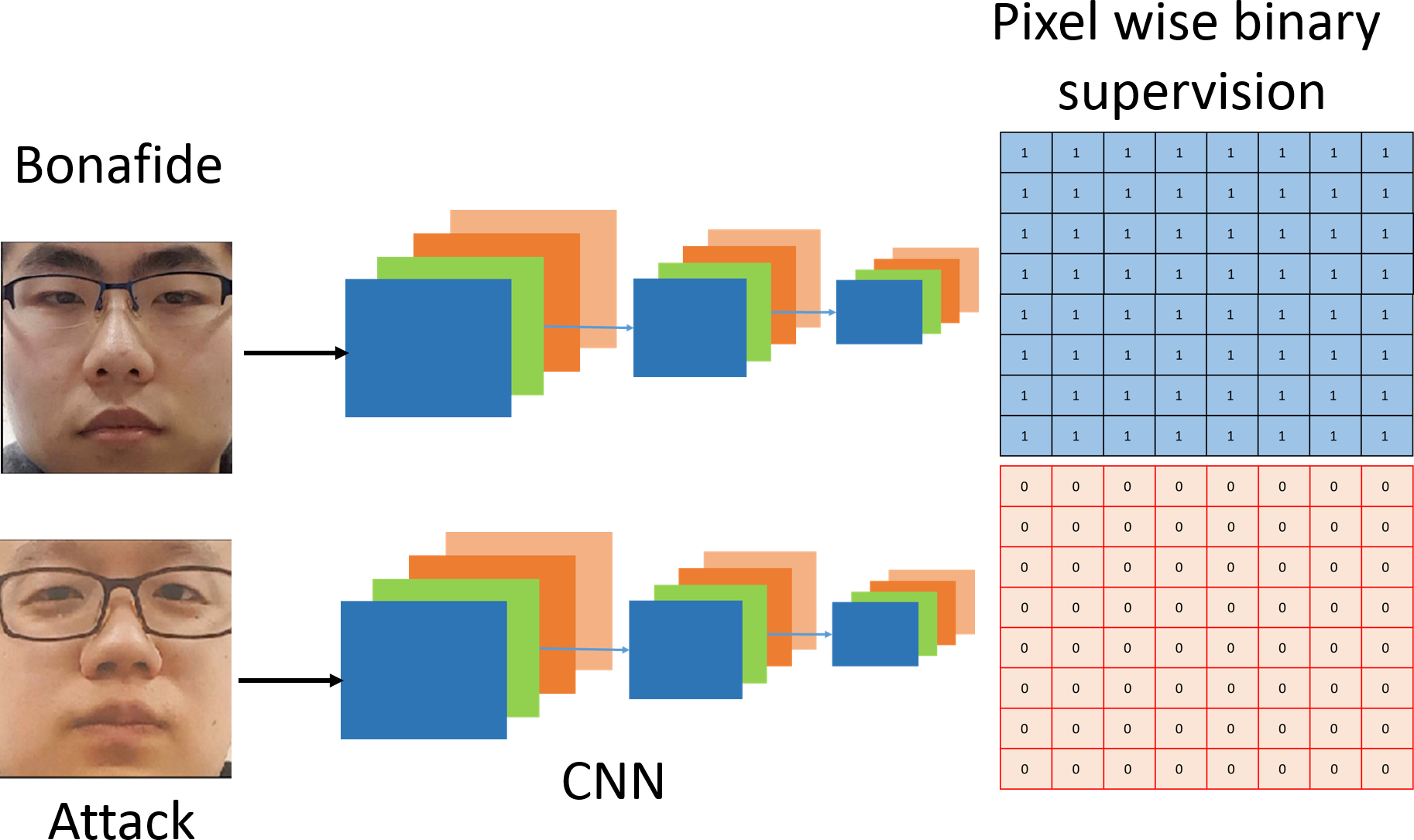}
        \caption{Figure showing the pixel-wise binary labels for \textit{bonafide} and attacks. Each pixel/patch is given a binary label depending on whether it is a \textit{bonafide} or an attack. In the testing phase, the mean of this feature map is used as the score.}\label{fig:pixel_wise_binary}

\end{figure} 

\begin{figure*}[t]
\centering
\includegraphics[width=0.98\textwidth]{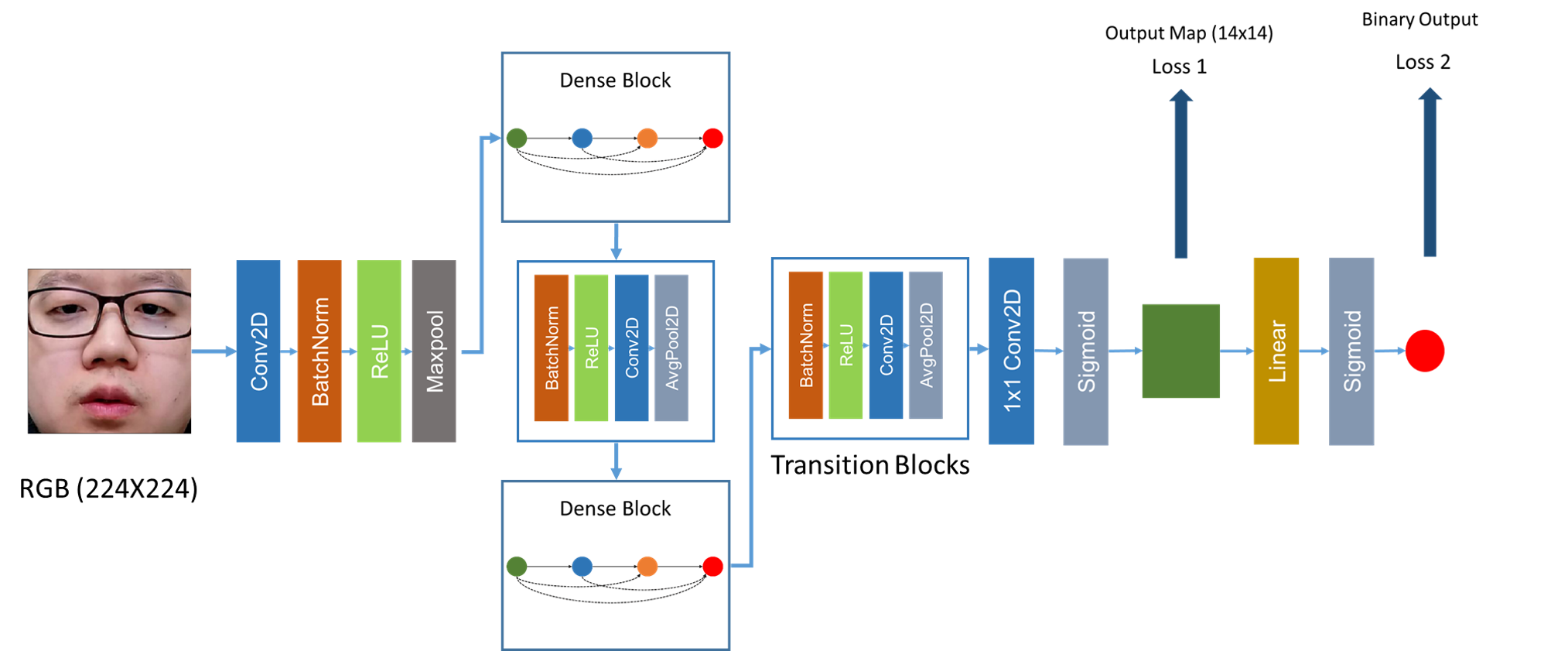}
\caption{ Diagram showing the proposed framework. Two outputs, i.e., a $14 \times 14$ feature map and a binary output are shown in the diagram. The dense blocks consist of multiple layers with each layer connected to all other layers. The feature maps are normalized and average pooled in the transition blocks.}
\label{fig:general_block}
\end{figure*}

This section introduces the proposed algorithm. A frame level CNN based framework which does not require temporal features for identifying presentation attacks is proposed in this section. The proposed framework uses a densely connected neural network trained using both binary and pixel-wise binary supervision (\textit{DeepPixBiS}). 

Recent papers \cite{atoum2017face}, \cite{liu2018learning} have shown that auxiliary supervision with depth and temporal information helps in achieving good performance in PAD task. However, using artificially synthesized depth maps as a proxy for PAD may not be ideal especially in frame level supervision. 

The depth supervision approaches try to generate the depth map of face for the \textit{bonafide} samples and a flat surface for the attacks. The idea is to learn to predict the true depth of the images presented in front of the biometric sensor. After training the network with the synthesized depth maps and flat masks, the network learns the subtle difference between \textit{bonafide} and attack images so that the correct depth map can be predicted. The predicted depth map is either used directly with $\ell2$ norm or with a classifier after extracting features from the depth map. It can be seen that the prediction of the accurate depth map is not essential for PAD task, rather predicting binary labels for pixels/patches would be enough.   

In this work, instead of using synthesized depth values for pixel-wise supervision, we use pixel-wise binary supervision. Both binary and pixel-wise binary supervision is used by adding a fully connected layer on top of the pixel-wise map.

This framework can be explained as follows. Considering a fully convolutional network, the output feature map from the network can be considered as the scores generated from the patches in an image, depending on the receptive fields of the convolutional filters in the network. Each pixel/ patch is labelled as \textit{bonafide} or attack as shown in Fig. \ref{fig:pixel_wise_binary}. In a way, this framework combines the advantages of patch-based methods and holistic CNN based methods using a single network. In the case of 2D attacks, we consider all patches have the same label. This obviates the requirement to compute the depth map while training models. This also makes it possible to extend the framework for partial attacks by annotating the ground truth mask regions. The advantage here is that the pixel-wise supervision forces the network to learn features which are shared, thus minimizing the number of learnable parameters significantly. 

Further, to combine the scores from the final feature map a fully connected layer is added on top of the final feature map. The loss function to minimize consists of the combination of both the binary loss and pixel-wise binary loss.

The details of the different parts of the proposed framework are detailed below.

\subsection{Preprocessing}
\label{subsec:preprocess}

In the first stage, face detection is carried out in the input images using MTCNN \cite{zhang2016joint} framework. Further, Supervised Descent Method (SDM) \cite{xiong2013supervised} is used to localize the facial landmark in the detected face region. The detected face image is aligned by making the eye centers horizontal. After alignment, the images are resized to a resolution of $224 \times 224$.

\subsection{Network architecture}

The proposed network is based on the DenseNet\cite{huang2017densely} architecture. The 
feature maps from multiple scales are used efficiently for PAD in this framework.


\subsubsection{DenseNet architecture}

The architecture used in this work is based on the DenseNet architecture proposed by Huang \textit{et al}. \cite{huang2017densely}. The main idea of DenseNet is to connect each layer to every other layer (with the same feature map size) in a feed-forward fashion. For each layer, feature maps from the previous layers are used as inputs. This implementation reduces the vanishing gradient problem as the dense connections introduce short paths from inputs to outputs. In each layer, feature maps are combined by concatenating previous feature maps. There are fewer parameters to train, and there is an improved flow of gradients to each layer. Another advantage of the DenseNet model is its implicit deep supervision, i.e., the individual layers receive supervision from the loss function because of the shorter connections. 

In our work, we reuse a pretrained model trained in the ImageNet dataset. The general block diagram of the framework is shown in Fig. \ref{fig:general_block}. First eight layers of the DenseNet \cite{huang2017densely} architecture are initialized from the pretrained weights. The layers selected consists of two dense blocks and two transition blocks. The dense blocks consist of dense connections between every layer with the same feature map size. The transition blocks normalize and downsample the feature maps. The output from the eight layers is of size  $14\times14$ with 384 channels. A $1\times1$ convolution layer is added along with sigmoid activation to produce the binary feature map. Further, a fully connected layer with sigmoid activation is added to produce the binary output. 

Binary Cross Entropy (BCE) is used as the loss function to train the model for both pixel-wise and binary output.

The equation for BCE for the pixel-wise loss is shown below.
\begin{equation}
\mathcal{L}_{pixel-wise-binary}=-{(y\log(p) + (1 - y)\log(1 - p))}
\end{equation}
where $y$ is the ground truth, ($y=0$ for attack and $y=1$ for \textit{bonafide}, for all values in the $14 \times 14$ feature map) and $p$ is predicted probability. 
The loss is averaged over pixels in the feature map.

Similarly $\mathcal{L}_{binary}$ is computed from the output using the binary label.
The loss to optimize is computed as the weighted sum of two losses:
\begin{equation}
\mathcal{L}=\lambda \mathcal{L}_{pixel-wise-binary}+ (1-\lambda)\mathcal{L}_{binary}
\end{equation}

We use the $\lambda$ value of 0.5 in the current implementation. Even though both losses are used in training, in the evaluation phase, only the pixel-wise map is used — the mean value of the map generated is used as the PA score in all the evaluations. 

\subsubsection{Implementation details}

The distribution of \textit{bonafide} and attacks were imbalanced in the training set. Class balancing was done by under-sampling the majority class. Data augmentation was performed during training using random horizontal flips with a probability of 0.5 along with random jitter in brightness, contrast, and saturation. The multi-task loss function is minimized using Adam Optimizer \cite{kingma2014adam}. A learning rate of $1\times10^{-4}$ was used with a weight decay parameter of  $1\times10^{-5}$. The mini-batch size used was 32, and the network was trained for 50 epochs on a GPU grid. While evaluating the framework, 20 frames were uniformly selected from each video, and the scores were averaged to compute the final PA score. The mean value of the $14 \times 14$ was used as the score for each frame in the video. The framework was implemented using PyTorch \cite{paszke2017automatic} library.

\section{Experiments}

\subsection{Databases and Evaluation Metrics}
\subsubsection{Databases}

Two recent databases, namely Replay-Mobile \cite{costa2016replay} and OULU-NPU \cite{boulkenafet2017oulu} are used in the experiments. The Replay-Mobile dataset consists of 1190 video clips of both photo and video attacks of 40 subjects under various lighting conditions. High-quality videos were recorded by iPad Mini2 and LG-G4. OULU-NPU is also a high-resolution dataset consisting of 4950 video clips. This database includes both video and photo attacks. The OULU-NPU dataset has four protocols each intended to test the generalization against variations in capturing conditions, attack devices, capturing devices and their combinations. We perform intra as well as cross-database testing in these two databases.


\subsubsection{Metrics}

In the OULU-NPU dataset, we use the recently standardized ISO/IEC 30107-3 metrics \cite{ISO} for our evaluation. We use Attack Presentation Classification Error Rate $APCER$, which corresponds to the worst error rate among the PAIs (print and video here), Bona Fide Presentation Classification Error Rate $BPCER$, which is the error rate in classifying a \textit{bonafide} as an attack, and $ACER$, which is computed as the mean of $APCER$ and $BPCER$:

\begin{equation}
ACER=\frac{APCER+BPCER}{2}.
\end{equation}

\begin{equation}
ACER=\frac{\max_{for PAI=1...C}(APCER_{PAI})+BPCER}{2}.
\end{equation}
Where $C$ is a PA category (print and video in OULU)

However, for cross-database testing, Half Total Error Rate (HTER) is adopted as done in previous literature \cite{liu2018learning}, which computes the average of False Rejection Rate (FRR) and the False Acceptance Rate (FAR):
\begin{equation}
HTER=\frac{FRR+FAR}{2}.
\end{equation}

The decision threshold is computed from the development set based on the equal error rate (EER) criterion. 

\subsection{Baseline systems}
We used two reproducible baselines available as open source in all the experiments. The first one
is based on the Image Quality Measures (IQM)~\cite{galbally2014image}. Each image is preprocessed
in a similar way as explained in Subsection \ref{subsec:preprocess}, and a 139-dimensional image quality feature vector is extracted. The extracted features are fed to an SVM, and the mean score of the frames is used as the final score. The second baseline uses the uniform Local Binary Patterns (LBP). After similar preprocessing, images are converted to gray-scale and 59 dimensional LBP histogram was computed. The resulting feature vector was used with an SVM classifier. These two baseline systems are denoted as \textit{IQM-SVM} and \textit{LBP-SVM} respectively. Apart from the baselines, best-performing methods from the public competition in OULU dataset \cite{boulkenafet2017competition} and recent methods are compared in the experimental section. 

\subsection{Intra testing}
We perform intra testing in both Replay Mobile and OULU datasets. 

\subsubsection{Intra testing in Replay Mobile dataset}

In the Replay Mobile dataset, intra testing is done with the `\textit{grandtest}' protocol. Scoring is performed video level by averaging the frame level scores. The comparison with the reproducible baselines is shown in Table \ref{tab:intra_rm}. 

\begin{figure}[ht]
\centering
\includegraphics[width=0.8\linewidth,page=2]{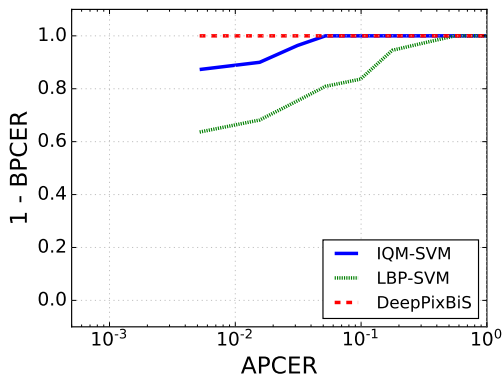}
\caption{ROCs for reproducible baselines and proposed \textit{DeepPixBiS} method in the \textit{eval} set of \textit{grandtest} protocol in Replay-Mobile dataset }
\label{fig:roc_baselines_RM}
\end{figure}

It can be seen that the proposed \textit{DeepPixBiS} method achieves 0\% HTER in the `\textit{grandtest}' protocol, outperforming all the baselines by a large margin. The ROC curves for the baselines and the proposed methods is shown in Fig. \ref{fig:roc_baselines_RM}.

\begin{table}[ht]
\begin{center}
\footnotesize
\begin{tabular}{c|c|c}
\toprule
  Method       & EER & HTER \\ \midrule
\textit{IQM-SVM}      &1.2 &3.9   \\ \hline
\textit{LBP-SVM}      &6.2	&12.1     \\ \hline
\textbf{DeepPixBiS}& 0.0    &0.0     \\ \bottomrule

\end{tabular}
\end{center}

\caption{Intra testing in Replay Mobile dataset}

\label{tab:intra_rm}

\end{table}


\subsubsection{Intra testing in OULU-NPU dataset}

For the OULU dataset, we follow a similar evaluation strategy as \cite{boulkenafet2017competition} for all four protocols. 

\begin{figure}[ht]
\centering
\includegraphics[width=0.8\linewidth,page=2]{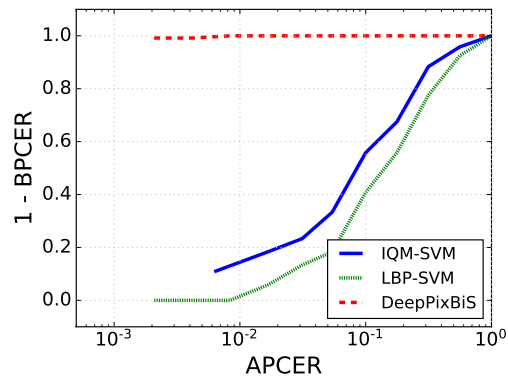}
\caption{ROCs for reproducible baselines and proposed \textit{DeepPixBiS} method in the \textit{eval} set of \textit{Protocol-1} in OULU-NPU dataset }
\label{fig:roc_baselines_OULU}
\end{figure}

The comparison with the methods taken from the papers, the proposed method, and our reproducible baselines are shown in Table \ref{tab:OULU}. From Table \ref{tab:OULU} it can be seen that the proposed method outperforms all the state of the art methods in `\textit{Protocol-1}'. The ROC curves for the baselines and the proposed methods for `\textit{Protocol-1}' is shown in Fig. \ref{fig:roc_baselines_OULU}.  The performance in `\textit{Protocol-4}' is the worst, which consists of unseen PAI and unseen environments. 

It was observed that for most of the cases the APCER was worse for print attacks as compared to video attacks. This could be because of the high quality of the prints in the OULU dataset. Motion-based methods are useful for improving the performance in case of print attacks. Fusion with such methods could improve the results at the cost of additional computational and time overhead.

\begin{table}[ht]
\begin{center}
\footnotesize
\resizebox{0.48\textwidth}{!}{
\begin{tabular}{c|c|c|c|c}
\toprule
Prot. & Method & APCER(\%) & BPCER(\%) & ACER(\%) \\
\midrule
\multirow{4}{*}{1} &CPqD &2.9 &10.8 & 6.9 \\
       \cline{2-5} &GRADIANT &1.3 &12.5 & 6.9 \\
       \cline{2-5} &FAS-BAS~\cite{liu2018learning} &1.6 &1.6 & 1.6 \\
       
       \cline{2-5} &\textit{IQM-SVM} &19.17  &30.83     &25\\
       \cline{2-5} &\textit{LBP-SVM} &12.92  &51.67  &32.29\\

       \cline{2-5} &\textbf{DeepPixBiS}  &0.83  &0  &0.42 \\
\midrule
\multirow{4}{*}{2} &MixedFASNet &9.7 &2.5 & 6.1 \\
       \cline{2-5} &FAS-BAS~\cite{liu2018learning} &2.7 &2.7 & 2.7 \\
       \cline{2-5} &GRADIANT &3.1 &1.9 & 2.5 \\
       \cline{2-5} &\textit{IQM-SVM} &12.5 &16.94 &14.72 \\
       \cline{2-5} &\textit{LBP-SVM} &30 &20.28 &25.14 \\

               \cline{2-5} &\textbf{DeepPixBiS} &11.39 &0.56 &5.97\\

\midrule
\multirow{4}{*}{3} &MixedFASNet &5.3$\pm$6.7 &7.8$\pm$5.5 &6.5$\pm$4.6 \\
       \cline{2-5} &GRADIANT &2.6$\pm$3.9 &5.0$\pm$5.3 &3.8$\pm$2.4 \\
       \cline{2-5} &FAS-BAS ~\cite{liu2018learning} &2.7$\pm$1.3 &3.1$\pm$1.7 &2.9$\pm$1.5 \\
       \cline{2-5} &\textit{IQM-SVM} &21.94$\pm$9.99  &21.95$\pm$16.79   &21.95$\pm$8.09
\\
       \cline{2-5} &\textit{LBP-SVM} &28.5$\pm$23.05 &23.33$\pm$17.98 &25.92$\pm$11.25 \\

    \cline{2-5} &\textbf{DeepPixBiS}  &11.67$\pm$19.57  &10.56$\pm$14.06   &11.11$\pm$9.4 \\

\midrule
\multirow{4}{*}{4} &Massy\_HNU &35.8$\pm$35.3 &8.3$\pm$4.1 &22.1$\pm$17.6 \\
       \cline{2-5} &GRADIANT &5.0$\pm$4.5 &15.0$\pm$7.1 &10.0$\pm$5.0 \\
       \cline{2-5} &FAS-BAS ~\cite{liu2018learning} &9.3$\pm$5.6 &10.4$\pm$6.0 &9.5$\pm$6.0 \\

       \cline{2-5} &\textit{IQM-SVM} &34.17$\pm$25.89  &39.17$\pm$23.35  &36.67$\pm$12.13 
 \\
       \cline{2-5} &\textit{LBP-SVM} &41.67$\pm$27.03  &55.0$\pm$21.21  &48.33$\pm$6.07 \\

    \cline{2-5} &\textbf{DeepPixBiS} &36.67$\pm$29.67  &13.33$\pm$16.75  &25.0$\pm$12.67\\

\bottomrule
\end{tabular}
}
\end{center}

\caption{The results of intra testing on four protocols of OULU-NPU~\cite{boulkenafet2017oulu}.} 
\label{tab:OULU}
\end{table}


\subsection{Cross database testing}
To test the generalization ability of the proposed framework, we perform cross-database experiments. Specifically, we do cross-database testing with the same datasets used in the intra testing experiments.

Many recently published papers report cross-database testing with CASIA-MFSD \cite{zhang2012face} and Replay-Attack \cite{chingovska2012effectiveness} databases, while reporting intra-dataset performance in OULU dataset. However, the best-performing methods in OULU are not evaluated in cross-database testings. This results in reporting over-optimistic results in intra testing as it is tuned for the specific dataset. To avoid this, we perform cross-testing with the exact same models used in the intra testing. In this way, the generalizability of the best-performing methods in a specific dataset can be examined in the cross-database testing scenario.

Two experiments were performed in cross-database testings. In the first one, the model trained on OULU `\textit{Protocol-1}' is tested on the `\textit{grandtest}' protocol of Replay Mobile dataset (OULU-RM). Conversely, in the second experiment, the model trained on the `\textit{grandtest}' protocol of Replay Mobile dataset is tested on the `\textit{Protocol-1}' of OULU dataset (RM-OULU).


The results of the cross-database testing are shown in Table \ref{tab:cross_test}. It can be seen that the model trained in OULU dataset achieves an HTER of 12.4\% in Replay Mobile dataset. The model used is the same as the one used in the intra-testing. It can be seen that the proposed method achieves much better generalization properties as compared to our reproducible baselines. 

While doing the cross-database testing in the reverse case, i.e., training on Replay Mobile and testing on OULU (RM-OULU), the HTER achieved is 22.7\%. Even though the performance of the proposed method is much better than the baselines, it can be seen that the generalization in RM-OULU testing is poor in general. This can be due to the limited amount of training data available for training in Replay Mobile dataset. Another reason could be the challenging nature of attacks in OULU. It is to be noted that the same model achieved nearly perfect separation in the intra testing scenario in Replay Mobile dataset.

\begin{table}[ht]

\begin{center}
\footnotesize

\begin{tabular}{@{}c|c|c|c|c@{}}
\toprule
\multirow{2}{*}{Method} & \multicolumn{2}{c|}{\begin{tabular}[c]{@{}c@{}}trained on\\ OULU\end{tabular}}                                     & \multicolumn{2}{c}{\begin{tabular}[c]{@{}c@{}}trained on\\ RM\end{tabular}}                                       \\ \cline{2-5}  
                        & \begin{tabular}[c]{@{}c@{}}tested on\\  OULU\end{tabular} & \begin{tabular}[c]{@{}c@{}}tested on\\ RM\end{tabular} & \begin{tabular}[c]{@{}c@{}}tested on\\ RM\end{tabular} & \begin{tabular}[c]{@{}c@{}}tested on \\ OULU\end{tabular} \\ \midrule
\textit{IQM-SVM}                                     & 24.6                                  & 31.6                                  & 3.9                                   & 42.3                                 \\ \hline
\textit{LBP-SVM}                                     & 32.2                                  & 35.0                                  & 12.1                                  & 43.6                                 \\ \hline
\textbf{DeepPixBiS}                                          & 0.4                                  & 12.4                                  & 0.0                                   & 22.7                                \\ \bottomrule
\end{tabular}
\end{center}

\caption{The results from the cross-database testing between OULU-NPU (`\textit{Protocol-1}') and Replay Mobile (`\textit{grandtest}' protocol)databases. HTER (\%) values are reported in the table.}
\label{tab:cross_test}
\end{table}

\subsection{Discussions}
From the experimental section, it can be seen that the proposed approach achieves perfect separation between \textit{bonafide} and attacks in Replay Mobile dataset and achieves good performance in protocols of OULU dataset. It is to be noted that the algorithm uses only frame level information for computing the scores. The cross-dataset experiments, especially OULU-RM shows good generalizability of the proposed approach across databases when sufficient training data is available. 

The main advantage of the proposed approach is its ease of implementation due to the frame level processing. 
The preprocessing part is simple including face detection and alignment. The cropped face image is fed to the trained CNN, and the output map is averaged to get the final PA score. A single forward pass through the CNN is enough for the PAD decision. This enables us to extend the framework by fusing other sources of information easily when computational and time overheads are not critical. Further, it is possible to extend the framework for partial attacks by modifying the ground truth binary masks. 


In general, one crucial limitation of CNN based methods for PAD is the limited amount of data available for training. Availability of a large amount of training data might improve the performance and generalization of the proposed approach further. 

\section{Conclusions and future directions}

\label{sec:conclusion}

In this work, we introduced a dense fully connected neural network architecture which was trained with pixel-wise binary supervision. The pixel-wise binary supervision on the output maps forces the network to learn shared representation utilizing the information from different patches. Unlike previous methods using multiple networks and ensembling of different models, here a single CNN model is used which can compute the PA score frame-wise. The proposed system only uses frame level information which makes it suitable for taking a decision quickly without the need for processing multiple frames which is useful in practical deployment scenarios. Further, the software to reproduce the system is made publicly available for fostering further extension of the work. From cross-database experiments, it can be seen that the performance is far from perfect. Fusion of multiple features has been shown to improve the accuracy at the cost of additional computational complexity. The proposed framework can be extended by adding temporal features to improve accuracy.  Availability of large scale databases for PAD might also improve the results from the proposed framework. 

\section*{Acknowledgment}

Part of this research is based upon work supported by the Office of the
Director of National Intelligence (ODNI), Intelligence Advanced Research
Projects Activity (IARPA), via IARPA R\&D Contract No. 2017-17020200005.
The views and conclusions contained herein are those of the authors and
should not be interpreted as necessarily representing the official
policies or endorsements, either expressed or implied, of the ODNI,
IARPA, or the U.S. Government. The U.S. Government is authorized to
reproduce and distribute reprints for Governmental purposes
notwithstanding any copyright annotation thereon.

{\small
\bibliographystyle{ieee}
\bibliography{egbib}
}

\end{document}